\documentclass[letterpaper,10pt,conference]{ieeeconf}
\IEEEoverridecommandlockouts
\usepackage{graphicx,booktabs,amsmath,amssymb,xcolor,tikz,url}
\usetikzlibrary{positioning,arrows.meta,fit}
\usepackage[hidelinks]{hyperref}
\title{\LARGE\bf SONIC-K1: Porting a Universal Humanoid Motion Tracker\\to a Small Humanoid on a Single GPU}

\author{Kyrylo Kolesnichenko$^{1,2}$, Irvin Steve Cardenas$^{2}$, and Jong-Hoon Kim$^{2}$%
\thanks{$^{1}$Institute of Social Sciences and Applied Informatics, Vilnius University, Lithuania}%
\thanks{$^{2}$Advanced Telerobotics Research Lab - Computer Science Department, Kent State University, Kent, OH 44242, USA
        {\tt\small kkolesni@kent.edu}}%
}

\begin{document}
\maketitle
\thispagestyle{empty}\pagestyle{empty}

\begin{abstract}
We present SONIC-K1, to our knowledge the first open-weights whole-body motion tracker for the Booster K1 humanoid,
obtained by adapting the SONIC framework to this smaller robot. The policy is trained from scratch on a single shared
GPU with a small fraction of the compute reported for SONIC. In MuJoCo simulation, it tracks nearly all held-out
motion clips and performs well on external benchmarks under SONIC's success criterion. The first release ran on the real K1 but shook its legs on steps and turns, and simulation showed a
sensitivity to actuation delay. We therefore post-trained it with the robot's deployment torque limits, randomized
0--40\,ms actuation delay and gain randomization. The post-trained v1.0 tracks 96.6\% of held-out clips under
real-robot torque limits and 40\,ms of delay, compared with 76.7\% before, and in our runs on the real K1 it walks and turns without the earlier leg shaking.
These results show broad motion-tracking capability on a modest training budget. Code, policy weights and evaluation tools are
publicly available.
\end{abstract}

\section{Introduction}\label{sec:intro}
\begin{figure}[t]\centering
\includegraphics[width=\columnwidth]{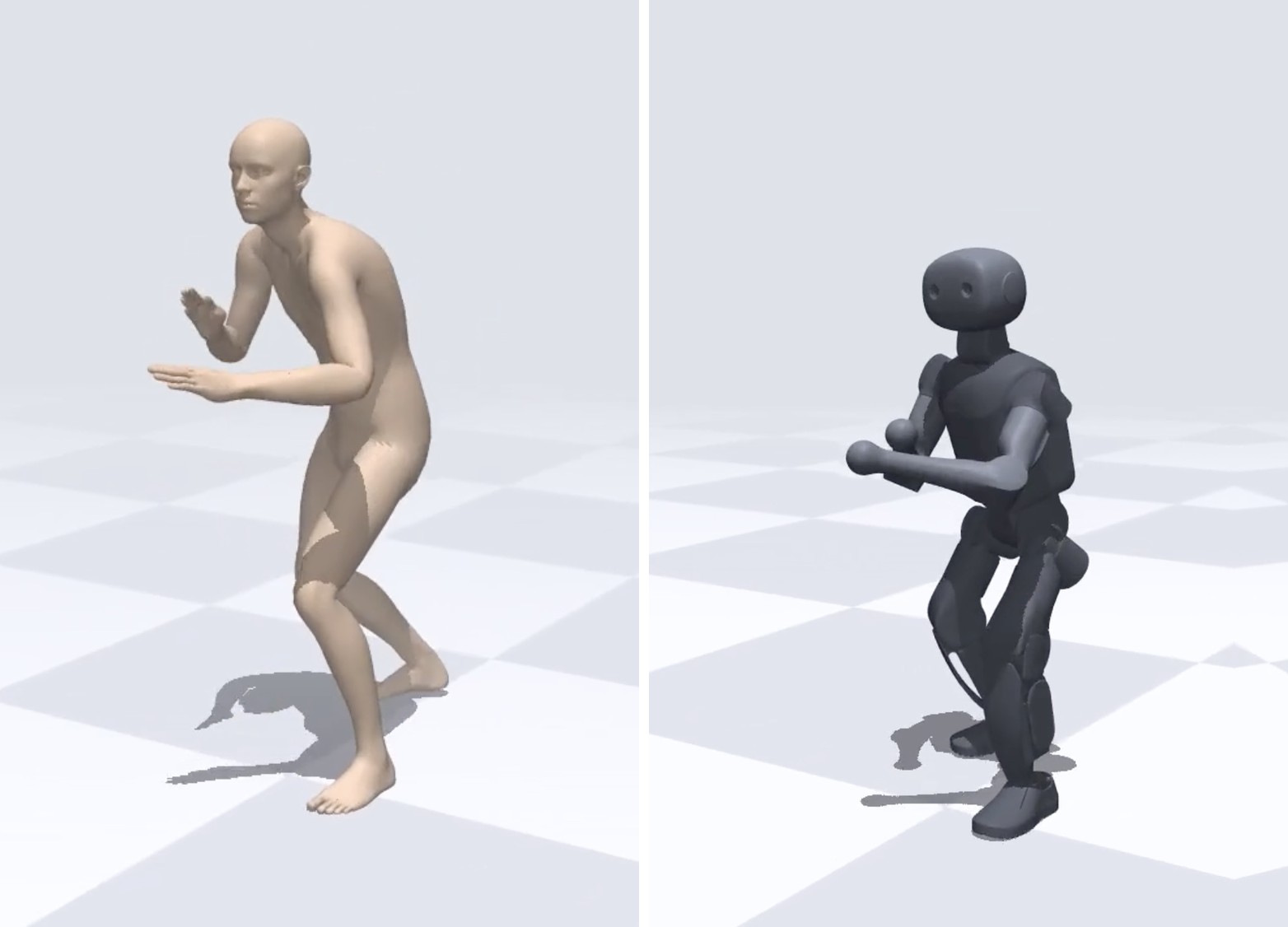}
\caption{A human reference motion (left) and SONIC-K1 tracking it on the Booster K1 in simulation (right).}\label{fig:teaser}
\end{figure}
A general motion tracker is a single policy that takes a reference trajectory and outputs joint commands. Once a robot has
one, motion capture, pose estimation from video and teleoperation can all drive the same
controller. SONIC~\cite{luo2025sonic} pushed
this approach to 700\,h of motion data and 128 GPUs. The code is open, but it is built for the Unitree G1, and most labs
cannot afford to train it at that scale.

Our lab works with the Booster K1~\cite{booster_k1}, a 22-DoF humanoid that is noticeably smaller than the G1, and we
wanted a SONIC-style tracker for it on the hardware we have. We kept SONIC's architecture, PPO settings and rewards,
retargeted 9,991 Bones-SEED clips~\cite{bones_seed} to the K1 with UMR~\cite{umr}, and trained from scratch for 50,000
iterations on a single H100 that was shared with other users, then post-trained it for 4,000 more iterations on one
RTX PRO 6000. The resulting model, SONIC-K1, is to our knowledge the
first general motion tracker for the K1 with public weights. It tracks 97.4\% of 1,985 held-out clips in MuJoCo~\cite{todorov2012mujoco} under SONIC's success rule. On the real robot this first release (v0.1)
shook its legs on steps and turns. A short post-training with the robot's torque limits and actuation delay (v1.0, 4,000
more iterations) fixed most of this: in simulation v1.0 tracks 96.6\% of held-out clips under real-robot limits and
40\,ms of delay (v0.1: 76.7\%), and in our robot runs it walks and turns without the earlier leg shaking.

This report makes three contributions:
\begin{enumerate}
\item SONIC-K1, the first open-weights general motion tracker for the Booster K1. It is trained with roughly 1--1.5\% of SONIC's compute, and we list every change we made to the original recipe.
\item An evaluation in two simulators on a 1,985-clip hold-out and on PHUMA,
LAFAN1 and 100STYLE, plus a robustness sweep.
\item A real-robot post-training recipe that makes the policy robust to the K1's torque limits and actuation delay,
tested on hardware, plus the released code (Apache-2.0), weights (CC BY-NC 4.0), evaluation tools and robot
application.
\end{enumerate}

\section{Related Work}\label{sec:related}
SONIC~\cite{luo2025sonic} trains a single tracker for the Unitree G1 on 700\,h of motion data. It uses per-modality
encoders, a quantized motion token and a shared decoder. We start from its public code and keep its architecture and
training recipe. To give a sense of scale for our results, we also list numbers that the SONIC paper reports for
BeyondMimic~\cite{liao2025beyondmimic} and Any2Track~\cite{zhang2025any2track}, two other trackers for the G1.

Our training data is the Bones-SEED motion-capture dataset~\cite{bones_seed}, which we map to the K1 with
UMR~\cite{umr}, a learned retargeting method for humanoid robots. Besides our own held-out Bones-SEED clips, we test on
PHUMA~\cite{phuma}, the external benchmark used in the SONIC paper, and on two standard human-motion datasets,
LAFAN1~\cite{harvey2020lafan} and 100STYLE~\cite{mason2022style100}.

Training runs in Isaac Lab~\cite{isaaclab} with PPO~\cite{schulman2017ppo}. We evaluate in
MuJoCo~\cite{todorov2012mujoco}, a different physics engine, to check that the policy does not depend on the training
simulator.

\section{Method: Porting SONIC to the K1}\label{sec:method}

\begin{figure*}[t]\centering
\begin{tikzpicture}[node distance=3mm and 4mm, every node/.style={font=\footnotesize},
  box/.style={draw, rounded corners=2pt, align=center, minimum height=9mm, inner sep=3pt, fill=#1},
  box/.default=white, arr/.style={-{Stealth[length=2mm]}, thick}]
\node[box=blue!8] (bones) {Bones-SEED\\142,220 clips};
\node[box=blue!8, right=of bones] (umr) {UMR retarget\\to K1};
\node[box=blue!8, right=of umr] (qa) {QA + feasibility\\62,427 clips};
\node[box=blue!8, right=of qa] (split) {train 9,991\\held-out 1,985};
\node[box=orange!12, right=of split] (train) {SONIC PPO, Isaac Lab\\1 shared H100: v0.1};
\node[box=green!12, right=of train] (rel) {Post-train, 1 RTX\\PRO 6000: v1.0};
\node[box=gray!12, right=of rel] (eval) {MuJoCo eval\\+ real K1};
\draw[arr] (bones)--(umr); \draw[arr] (umr)--(qa); \draw[arr] (qa)--(split); \draw[arr] (split)--(train);
\draw[arr] (train)--(rel); \draw[arr] (rel)--(eval);
\end{tikzpicture}
\caption{SONIC-K1 pipeline. Motion data is not redistributed; only the policy and code are released.}\label{fig:pipeline}
\end{figure*}
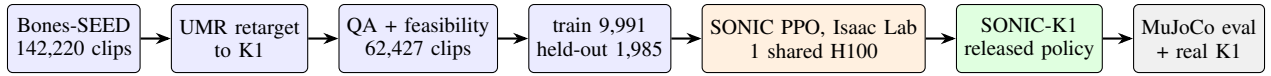

\subsection{Robot and data}
The K1 has 22 actuated joints: 2 in the head, 4 in each arm and 6 in each leg. It has no waist and no wrists. We use
Booster's official \texttt{K1\_22dof} model for both retargeting and training. Bones-SEED contains 142,220
clips, all of which we retargeted to the K1 with UMR. An automatic check then removes clips with
joint-limit violations, feet going through the floor, velocity spikes or NaNs, and tags each clip as feasible or not for the
K1 (for example, motions that need a waist are not). We keep only clips that pass the check and are tagged feasible,
which leaves 62,427. We trained on 9,991 of them and kept 1,985 aside as a test
set, chosen before training started and with no overlap (Fig.~\ref{fig:pipeline}).

\subsection{Policy and training}
The architecture is unchanged from SONIC. Encoders turn a reference motion into a 64-D quantized
(FSQ) token, and an MLP decoder combines the token with proprioception to produce joint targets. A PD
controller tracks those targets; the policy runs at 50\,Hz and physics at 200\,Hz. The actor has 25.1M parameters. We
trained the robot-motion and teleoperation encoders. The SMPL encoder is in the network but never saw any data.

PPO hyper-parameters, adaptive motion sampling and rewards are the same as in SONIC, and Table~\ref{tab:diff} lists what
we changed. The biggest difference is scale. We ran 16,384 environments for 50,000 iterations on one H100 NVL, which took
135.4\,h of wall-clock time because other jobs were sharing the GPU (about 103\,h if it had been ours alone). That is
roughly 1--1.5\% of SONIC's 128 GPUs $\times$ $\sim$3 days. The real-robot post-training (v1.0,
Sec.~\ref{sec:posttrain}) added 4,000 iterations, about 9.2\,h on one NVIDIA RTX PRO 6000. We also raised the initial action standard deviation from
0.05 to 0.5, because with 0.05 the KL-adaptive scheme throttled the learning rate on the K1. Finally, we added a reward
for tracking the head joints, since the K1 head is actuated, and scaled pushes and center-of-mass randomization by 0.7
for the smaller body. v0.1 did not model actuation delay.

\begin{table}[t]\centering\caption{SONIC vs.\ SONIC-K1.}\label{tab:diff}
\footnotesize
\begin{tabular}{@{}lll@{}}\toprule
& SONIC & SONIC-K1 (ours) \\\midrule
Robot & Unitree G1, 29 DoF & Booster K1, 22 DoF \\
Training data & 700 h in-house mocap & 9,991 Bones-SEED clips \\
Compute & 128 GPUs $\times$ $\sim$3 days & 1 shared H100, 135.4 h \\
Post-training & -- & v1.0: 1 RTX PRO 6000, 9.2 h \\
Parallel envs & $\sim$524K & 16,384 \\
Iterations & $\sim$100K & 50,000 (+4,000, v1.0) \\
Init. action std & 0.05 & 0.5 \\
Rewards & SONIC Table 3 & + head joints; v1.0: + torque \\
Delay & none & 0--40 ms (v1.0) \\
\bottomrule\end{tabular}\end{table}

\subsection{Evaluation protocol}
We use SONIC's success rule: a rollout fails if the root or any end-effector ends up more than 0.25\,m away in height
from the reference. We also report two other rules. \emph{No-fall} only checks that the robot stays up
(root height within 0.25\,m, orientation within 1\,rad). \emph{Official}, the stricter rule in the SONIC repository,
also limits world-frame foot position to 0.2\,m. Tracking error is MPJPE-L over 12 K1 links, in mm, averaged over
successful clips. For
the MuJoCo runs we use the robot-motion encoder and deterministic actions.

\section{Experiments}\label{sec:exp}
All results in this section are for the first release (v0.1, 50,000 iterations), use 10\,s windows, and apply
SONIC's 0.25\,m success rule. Section~\ref{sec:posttrain} covers the real-robot post-training (v1.0).

\begin{table*}[t]\centering\caption{SONIC-K1 v0.1 results in MuJoCo. Success in \%, MPJPE-L in mm. The G1 columns are numbers reported in the SONIC paper~\cite{luo2025sonic}, shown for scale only.}\label{tab:bench}
\begin{tabular}{@{}lr|rrr|r|rrr@{}}\toprule
& & \multicolumn{3}{c|}{SONIC-K1 success} & & \multicolumn{3}{c}{G1, reported (SONIC rule)}\\
Test set & Clips & SONIC rule & No-fall & Official & MPJPE-L & SONIC & BeyondMimic & Any2Track \\\midrule
Bones-SEED held-out & 1,985 & \textbf{97.4} & 98.4 & 91.9 & 15.4 & 98.5 & 82.0 & 61.6 \\
PHUMA & 1,758 & \textbf{93.6} & 97.3 & 85.0 & 22.4 & 97.2 & 73.8 & 78.5 \\
LAFAN1 (10\,s windows) & 1,390 & \textbf{70.6} & 75.5 & 45.0 & 31.9 & -- & -- & -- \\
LAFAN1 (full clips) & 65 & \textbf{18.5} & 27.7 & 0.0 & 28.1 & -- & -- & -- \\
100STYLE & 200 & \textbf{99.0} & 98.5 & 77.0 & 22.5 & -- & -- & -- \\
\bottomrule\end{tabular}\end{table*}

\subsection{Benchmarks}\label{sec:bench}
Table~\ref{tab:bench} shows the MuJoCo results. On the 1,985 held-out Bones-SEED clips the policy succeeds 97.4\% of the
time, with an MPJPE-L of 15.4\,mm. PHUMA provides G1 trajectories, so we sampled 2,000 motions across its 15
sources and mapped them from the G1 to the K1 with our own inverse kinematics. Of these, 1,758 passed our checks, and the
policy succeeds on 93.6\%. On 100STYLE we used the forward walk and run of all 100 styles, and
the policy succeeds on 99.0\%. LAFAN1 was the hardest set, at 70.6\% of 10\,s windows; we come back
to it in the limitations.

The right side of Table~\ref{tab:bench} lists G1 numbers for SONIC, BeyondMimic and Any2Track, copied
from the SONIC paper. We did not rerun these methods. The
robot, data, retargeting and test clips are all different, so these columns are only there to give a sense of scale and
should not be read as a head-to-head comparison.

We also checked the same hold-out in Isaac Lab, the simulator we trained in. No-fall success is 98.7\% there and 98.4\%
in MuJoCo, and official-rule success in Isaac Lab is 93.1\%, so moving between simulators costs very little. The amount
of training data makes a large difference. An earlier run trained on only 10 clips reached 63.6\% official success on
the same hold-out (33.5\,mm MPJPE-L). The released 9,991-clip model reaches 93.1\% (14.7\,mm).

\subsection{Robustness}\label{sec:robust}
\begin{figure}[t]\centering
\includegraphics[width=\columnwidth]{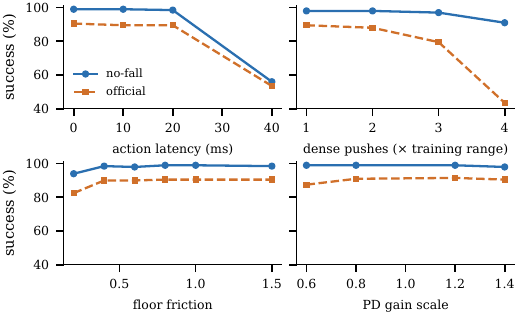}
\caption{Perturbation sweep for SONIC-K1 v0.1 in MuJoCo (200 held-out clips per setting). Latency is the only perturbation
that causes a large drop.}\label{fig:robust}
\end{figure}

Fig.~\ref{fig:robust} shows a perturbation sweep in MuJoCo with 200 held-out clips per setting, reported under the
no-fall and official rules. No-fall success stays at 98.0\% or higher with PD gains scaled anywhere from 0.6 to
1.4$\times$, with 70\% motor strength, with a 2\,kg payload, with 3$\times$ observation noise and with Booster's torque
caps. Dense pushes up to 3$\times$ the training range still give 97.0\%, and very low friction (0.2) costs about 5
points. Latency is the exception. Success is 98.5\% at 20\,ms of delay but falls to 56.0\% at 40\,ms.

\subsection{First test on the real K1}\label{sec:real}
We exported the policy to ONNX and ran it at 50\,Hz on the robot's onboard computer, with the robot on a safety belt,
using the sample clips from the release kit. Upper-body clips, calm motions and slower whole-body motions such as squats
ran smoothly. Clips with steps and turns were less reliable, with visible jitter in the legs. We do not have motion capture for the robot, so we cannot report tracking error on hardware.
Section~\ref{sec:posttrain} addresses the leg problem.

\section{Post-Training for the Real K1 (v1.0)}\label{sec:posttrain}
\textbf{Problem.} On the real robot the first release (v0.1) shook its legs on steps and turns
(Sec.~\ref{sec:real}). Two facts point to the training plant. First, training allowed leg torques of
68/76/38/112/38\,Nm for hip pitch/roll/yaw, knee and ankle, while Booster's deployment limits are 30/35/20/40/20\,Nm.
Second, training had no actuation delay, and in simulation v0.1 drops from 98.5\% to 56.0\% success when the delay grows
from 20 to 40\,ms (Sec.~\ref{sec:robust}). Raising $k_d$ helped in simulation but made the robot worse, and fine-tuning
on a single failing clip did not help on hardware either.

\textbf{Method.} We resume v0.1 with four changes and keep observations, rewards, PD gains, action scales and data
unchanged, so the checkpoint stays valid:
\begin{itemize}
\item \emph{Deployment torque limits} on hips, knees and ankles (30/35/20/40/20\,Nm); arm and head limits unchanged.
\item \emph{Actuation delay}: in every episode each joint command is delayed by a uniform 0--8 physics steps
(0--40\,ms). The upper bound rises linearly from 0 to 40\,ms over the first 2,000 iterations.
\item \emph{Gain randomization}: $k_p$ and $k_d$ scaled by $\mathcal{U}(0.9,1.1)$ per episode for all joints, as in
PBHC~\cite{xie2025kungfubot}.
\item \emph{Torque-near-limit penalty}, weight $-0.5$:
$\sum_{j\in\text{legs}} \mathrm{clamp}\big(|\tau^{\text{PD}}_j|/\tau^{\max}_j - 0.9,\,0,\,1\big)$ on the unclipped PD demand.
\end{itemize}
The run used 16,384 environments on one NVIDIA RTX PRO 6000 (96\,GB) at 8.3\,s per iteration. The released v1.0 is
iteration 54,000, i.e.\ 4,000 iterations or about 9.2\,GPU-hours on top of v0.1. Held-out metrics were flat within
evaluation noise from about 53,000 iterations on. One practical point: SONIC's trainer keeps a 1,024-clip window on the
GPU and redraws it every 250 iterations, which on a single GPU made the training metrics jump at every redraw. We
therefore keep all 9,991 clips on the GPU.

\begin{table}[t]\centering\caption{Real-robot post-training: v0.1 (50,000 iterations) vs.\ v1.0 (54,000). Success in \% under SONIC's rule (Isaac rows: official rule, full-length clips, Booster torque caps), MPJPE-L in mm over successful clips. Real-robot = Booster torque caps + observation noise + the given delay. Both checkpoints evaluated with the same code and settings.}\label{tab:v1}
\footnotesize\setlength{\tabcolsep}{3pt}
\begin{tabular}{@{}llr|rr|rr@{}}\toprule
& & & \multicolumn{2}{c|}{v0.1} & \multicolumn{2}{c}{v1.0}\\
Test set & Setting & Clips & succ & L & succ & L\\\midrule
MuJoCo held-out & nominal & 1,985 & 97.5 & 15.4 & 97.4 & 17.1\\
 & real-robot 20 ms & 1,985 & 95.5 & 18.1 & 97.1 & 17.6\\
 & real-robot 40 ms & 1,985 & 76.7 & 21.2 & 96.6 & 18.2\\
MuJoCo seen & real-robot 40 ms & 2,000 & 78.3 & 22.5 & 97.9 & 18.1\\
PHUMA & nominal & 1,758 & 93.8 & 22.6 & 92.3 & 25.2\\
 & real-robot 40 ms & 1,758 & 78.0 & 23.9 & 91.7 & 25.4\\
Isaac official & no delay & 1,985 & 92.3 & 15.3 & 90.7 & 17.0\\
 & 40 ms delay & 1,985 & 70.4 & 16.1 & 89.4 & 16.9\\
\bottomrule\end{tabular}\end{table}

\textbf{Simulation results.} Table~\ref{tab:v1} compares both versions with the same evaluation code. Under the
real-robot torque limits with 40\,ms of delay, v1.0 tracks 96.6\% of the 1,985 held-out clips instead of 76.7\%, and
PHUMA improves from 78.0\% to 91.7\%. In Isaac Lab under the stricter official rule, success at 40\,ms rises from 70.4\%
to 89.4\%. Without delay, v1.0 stays close to v0.1 (97.4\% vs.\ 97.5\% on the held-out set), at slightly lower
accuracy: MPJPE-L 17.1 vs.\ 15.4\,mm, and 92.3\% vs.\ 93.8\% on PHUMA.

\begin{table}[t]\centering\caption{Real K1 (belt, kd$\times$1.0): v0.1 vs.\ the post-training run at 53,000 iterations on the same clips (first run of each). Leg jitter = RMS of 0.1\,s high-passed leg joint angles; capped = control ticks with a torque cap active; HF = share of leg joint-velocity power above 4\,Hz. Lunge compared over the first 7.1\,s of both runs.}\label{tab:robot}
\footnotesize\setlength{\tabcolsep}{3pt}
\begin{tabular}{@{}lcccc@{}}\toprule
Clip & jitter (deg) & gyro HF (rad/s) & capped (\%) & HF (\%)\\\midrule
yawn & 0.038 $\to$ 0.032 & 0.023 $\to$ 0.015 & 0.0 $\to$ 0.0 & 44 $\to$ 49\\
squat & 0.202 $\to$ 0.140 & 0.088 $\to$ 0.064 & 1.5 $\to$ 0.0 & 5 $\to$ 3\\
forward lunge & 0.597 $\to$ 0.365 & 0.226 $\to$ 0.112 & 8.1 $\to$ 1.1 & 29 $\to$ 10\\
\bottomrule\end{tabular}\end{table}

\textbf{Real robot.} We ran v1.0 on the K1 with the same setup as before. In our runs (observed and on video), walking,
turning, squats and lunges no longer showed the leg shaking, including the walk-and-turn clip on which v0.1 shook.
Table~\ref{tab:robot} gives leg jitter measured on three clips from robot logs of the same run at 53,000 iterations,
which matches v1.0 within noise in simulation. On the forward lunge, leg jitter falls
from 0.597 to 0.365\,deg, the share of control ticks at a torque limit from 8.1\% to 1.1\%, and the share of leg
joint-velocity power above 4\,Hz from 29\% to 10\%. We release the robot application with v1.0: a motion menu on the
Booster SDK with a damping key that works at any time and guards for tilt, falls, joint speed and stale sensor data.

\section{Weaknesses and Limitations}\label{sec:limits}
Post-training largely removed the leg shaking we observed on the real robot; on some motions the hands sometimes touch the body. Motions
that need floor contact or props, such as falling and getting up in LAFAN1, are outside our training data, and very long
clips build up global drift. Most of our evaluation is in simulation, with one training seed, and we did not ablate the four post-training changes. The SMPL and video encoders
are not trained yet, so video input is retargeted to the K1 first.

\section{Conclusion and Future Work}\label{sec:conclusion}
SONIC-K1 is the first open-weights general motion tracker for the Booster K1. SONIC was designed for 128 GPUs, but our
port, trained on one GPU, still reaches 97.4\% success on held-out motion under SONIC's own rule. A short post-training with the robot's torque limits, actuation delay and gain randomization made it work on
the real K1: v1.0 tracks 96.6\% of held-out clips under real-robot limits and 40\,ms of delay, and in our runs it walks and turns without the earlier leg shaking
on hardware.

Our next step is a proper evaluation on the robot with tracking metrics. Further plans are a larger model trained on the full 62K-clip
pool, whole-body teleoperation through the teleoperation encoder we already trained, and driving the robot from video.
For video we currently estimate human pose and retarget it, and a trained SMPL encoder would remove the retargeting
step.

\textbf{Release.} The code, including the robot application, is available under Apache-2.0 at \url{https://github.com/Kyk0/sonic-k1}, the weights (v1.0 and v0.1) under
CC BY-NC 4.0 at \url{https://huggingface.co/Kyry10/sonic-k1}, and the project page is \url{https://kyry.me/sonic-k1}.
The Booster K1 model is BSD-3-Clause. Training data includes Motion Data by Bones Studio (\url{https://bones.studio/}).
We do not redistribute the motion data, either raw or retargeted.

\section*{Acknowledgments}
We thank NVIDIA GEAR for open-sourcing SONIC, Bones Studio for Bones-SEED, the UMR authors, and Booster Robotics for the
K1 models.

\bibliographystyle{IEEEtran}
\bibliography{refs}

\begin{thebibliography}{10}
\providecommand{\url}[1]{#1}
\csname url@samestyle\endcsname
\providecommand{\newblock}{\relax}
\providecommand{\bibinfo}[2]{#2}
\providecommand{\BIBentrySTDinterwordspacing}{\spaceskip=0pt\relax}
\providecommand{\BIBentryALTinterwordstretchfactor}{4}
\providecommand{\BIBentryALTinterwordspacing}{\spaceskip=\fontdimen2\font plus
\BIBentryALTinterwordstretchfactor\fontdimen3\font minus
  \fontdimen4\font\relax}
\providecommand{\BIBforeignlanguage}[2]{{%
\expandafter\ifx\csname l@#1\endcsname\relax
\typeout{** WARNING: IEEEtran.bst: No hyphenation pattern has been}%
\typeout{** loaded for the language `#1'. Using the pattern for}%
\typeout{** the default language instead.}%
\else
\language=\csname l@#1\endcsname
\fi
#2}}
\providecommand{\BIBdecl}{\relax}
\BIBdecl

\bibitem{luo2025sonic}
Z.~Luo, Y.~Yuan, T.~Wang, C.~Li, F.~Casta{\~n}eda, S.~Chen, Z.-A. Cao, J.~Li,
  D.~Minor, Q.~Ben, J.~Park, D.~Sami, Z.~Wang, X.~Da, R.~Ding, C.~Hogg,
  L.~Song, E.~Lim, E.~Jeong, T.~He, H.~Xue, W.~Xiao, S.~Yuen, J.~Kautz,
  Y.~Chang, U.~Iqbal, L.~Fan, and Y.~Zhu, ``{SONIC}: Supersizing motion
  tracking for natural humanoid whole-body control,'' \emph{arXiv preprint
  arXiv:2511.07820}, 2025.

\bibitem{booster_k1}
{Booster Robotics}, ``Booster {K1} humanoid robot,''
  \url{https://www.booster.tech/}.

\bibitem{bones_seed}
{Bones Studio}, ``{Bones-SEED} motion dataset,''
  \url{https://huggingface.co/datasets/bones-studio/seed},
  \url{https://bones.studio/}.

\bibitem{umr}
H.~Cao, Y.~Fang, T.~Kwon, R.~Yu, J.~Ma, J.~Tan, Y.~Zhou, B.~Du, Y.~Gu, Y.~Gao,
  R.~Dai, L.~Han, and R.~Xu, ``Unified motion retargeting for humanoids with
  learned point cloud correspondence,'' \emph{arXiv preprint arXiv:2609.02134},
  2026.

\bibitem{todorov2012mujoco}
E.~Todorov, T.~Erez, and Y.~Tassa, ``{MuJoCo}: A physics engine for model-based
  control,'' in \emph{2012 IEEE/RSJ International Conference on Intelligent
  Robots and Systems (IROS)}, 2012, pp. 5026--5033.

\bibitem{liao2025beyondmimic}
Q.~Liao, T.~E. Truong, X.~Huang, Y.~Gao, G.~Tevet, K.~Sreenath, and C.~K. Liu,
  ``{BeyondMimic}: From motion tracking to versatile humanoid control via
  guided diffusion,'' \emph{arXiv preprint arXiv:2508.08241}, 2025.

\bibitem{zhang2025any2track}
Z.~Zhang, J.~Guo, C.~Chen, J.~Wang, C.~Lin, Y.~Lian, H.~Xue, Z.~Wang, M.~Liu,
  J.~Lyu, H.~Liu, H.~Wang, and L.~Yi, ``Track any motions under any
  disturbances,'' \emph{arXiv preprint arXiv:2509.13833}, 2025.

\bibitem{phuma}
K.~Lee, S.~Kim, Y.~Lee, M.~Park, H.~Kim, D.~Hwang, D.~Kim, H.~Lee, and J.~Choo,
  ``{PHUMA}: Physically reliable humanoid locomotion dataset,'' \emph{arXiv
  preprint arXiv:2510.26236}, 2025.

\bibitem{harvey2020lafan}
F.~G. Harvey, M.~Yurick, D.~Nowrouzezahrai, and C.~Pal, ``Robust motion
  in-betweening,'' \emph{ACM Transactions on Graphics}, vol.~39, no.~4, 2020.

\bibitem{mason2022style100}
I.~Mason, S.~Starke, and T.~Komura, ``Real-time style modelling of human
  locomotion via feature-wise transformations and local motion phases,''
  \emph{Proceedings of the ACM on Computer Graphics and Interactive
  Techniques}, vol.~5, no.~1, pp. 1--18, 2022.

\bibitem{isaaclab}
{NVIDIA}, M.~Mittal, P.~Roth, J.~Tigue, A.~Richard, O.~Zhang, P.~Du,
  A.~Serrano-Mu{\~n}oz, X.~Yao, R.~Zurbr{\"u}gg, N.~Rudin \emph{et~al.},
  ``{Isaac Lab}: A {GPU}-accelerated simulation framework for multi-modal robot
  learning,'' \emph{arXiv preprint arXiv:2511.04831}, 2025.

\bibitem{schulman2017ppo}
J.~Schulman, F.~Wolski, P.~Dhariwal, A.~Radford, and O.~Klimov, ``Proximal
  policy optimization algorithms,'' \emph{arXiv preprint arXiv:1707.06347},
  2017.

\bibitem{xie2025kungfubot}
W.~Xie, J.~Han, J.~Zheng, H.~Li, X.~Liu, J.~Shi, W.~Zhang, C.~Bai, and X.~Li,
  ``{KungfuBot}: Physics-based humanoid whole-body control for learning
  highly-dynamic skills,'' \emph{arXiv preprint arXiv:2506.12851}, 2025.

\end{thebibliography}

\end{document}